\documentclass[letterpaper, 10 pt, conference]{ieeeconf}  
\usepackage{graphicx}
\usepackage[hidelinks]{hyperref}
\usepackage{amsmath}
\usepackage{amssymb}
\usepackage{subcaption}
\usepackage{cite}
\usepackage{booktabs}  
\usepackage{siunitx}   
\usepackage[font=small]{caption}
\usepackage{comment}
\IEEEoverridecommandlockouts                              

\usepackage{comment}
\usepackage{xcolor}

\definecolor{orange}{RGB}{224, 71, 0}

\usepackage[normalem]{ulem}

\title{\LARGE \bf
    Trajectory Optimization for Self-Wrap-Aware Cable-Towed Planar Object Manipulation under Implicit Tension Constraints
}

 \author{Yu Li$^{1}$, Amin Fakhari$^{2}$ and Hamid Sadeghian$^{1}$
  \thanks{$^{1}$ Munich Institute of Robotics and Machine Intelligence, Technical University of Munich, Germany. Corresponding Author's e-mail: yu.li@tum.de
  $^{2}$ The author is with the Department of Mechanical Engineering, Stony Brook University, Stony Brook, NY 11794, USA.
  }
  \thanks{
The authors would like to thank the Federal Ministry of Research, Technology, and Space (BMFTR) for its support as part of the research program Communication Systems “Souverän. Digital. Vernetzt.”. Joint project 6G-life, project identification number: 16KIS2414.
This work has also received funding from the European Union’s Horizon Europe research and innovation programme as part of the project FlexCycle under grant agreement No. 101189600.
  }
}

\begin{document}

\maketitle
\thispagestyle{empty}
\pagestyle{empty}


\begin{abstract}
Cable/rope elements are pervasive in deformable-object manipulation, often serving as a deformable force-transmission medium whose routing and contact determine how wrenches are delivered. In cable-towed manipulation, transmission is unilateral and hybrid: the tether can pull only when taut and becomes force-free when slack; in practice, the tether may also contact the object boundary and self-wrap around edges, which is not merely collision avoidance but a change of the wrench transmission channel by shifting the effective application point and moment arm, thereby coupling routing geometry with rigid-body motion and tensioning. We formulate self-wrap towing as a routing-aware, tensioning-implicit trajectory optimization (TITO) problem that couples (i) a tensioning-implicit taut/slack constraint and (ii) routing-conditioned transmission maps for effective length and wrench, and we build a relaxation hierarchy from a strict mode-conditioned reference to three tractable relaxations: Full-Mode Relaxation (FMR), Binary-Mode Relaxation (BMR), and Implicit-Mode Relaxation (IMR). Across planar towing tasks, we find that making routing an explicit decision often yields conservative solutions that stay near switching boundaries, whereas IMR induces self-wrap through state evolution and exploits the redirected torque channel whenever turning requires it.
\end{abstract}

\section{Introduction}

Cable/rope elements are central to deformable-object manipulation because routing and contact determine how wrenches are transmitted. In cable-towed manipulation, the cable is primarily a transmission medium rather than the manipulated object, yet its unilateral physics yields hybrid dynamics: the tether can only pull, producing tension when taut and zero force when slack. This taut/slack switching is commonly expressed via complementarity and appears in both ground towing and suspended-load systems \cite{yang2022ral_cabletowed,sreenath2013icra_hybridload,cruz2017autonrobots}. Prior work has developed planning/control pipelines under a simplifying geometry where the tether remains a straight segment to a fixed attachment point \cite{yang2022ral_cabletowed,tang2015miqp}, sometimes reinforced by angular restrictions to simplify feasible-set computation and mode handling \cite{yang2022ral_cabletowed}.

\begin{figure}[t]
    \centering
    \includegraphics[width=\linewidth]{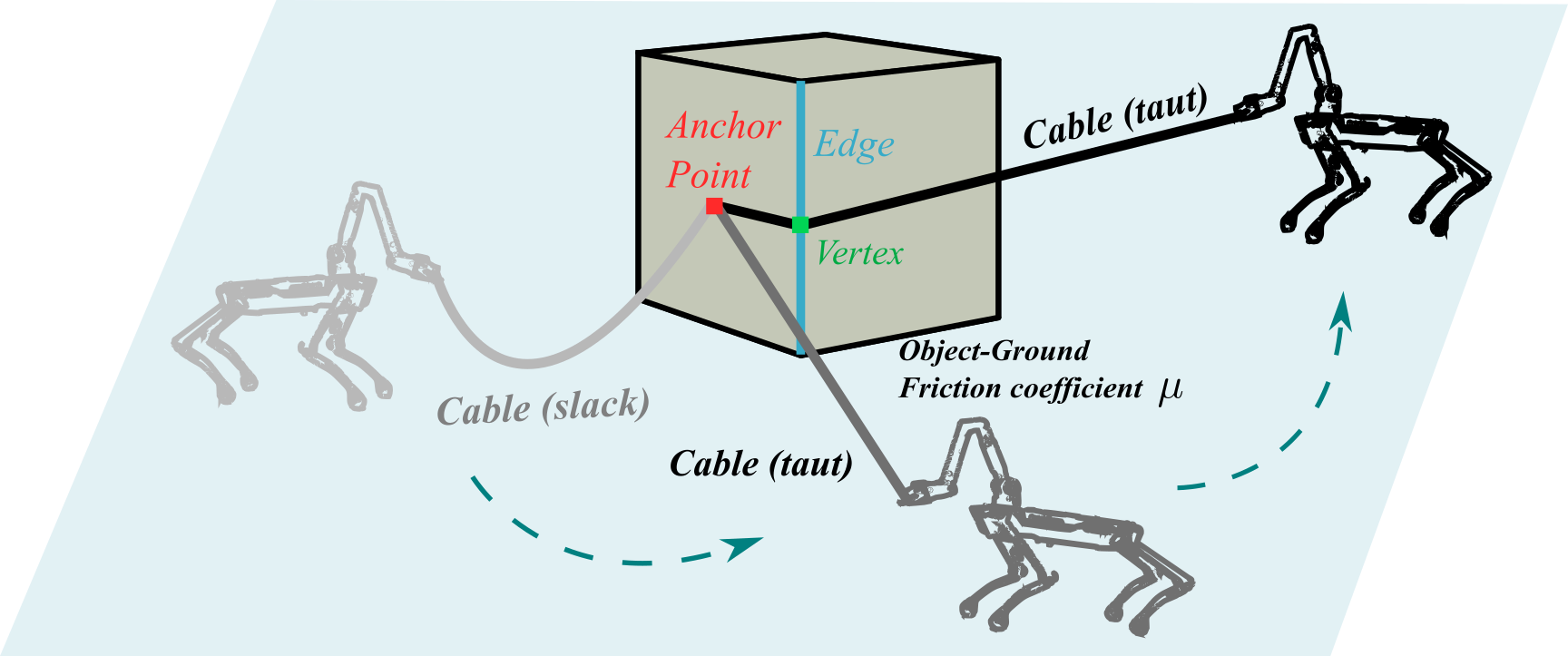}
    \caption{Illustration of cable-towed manipulation of a box by a mobile robot at three phases: (i) slack cable, (ii) taut cable without self-wrap contact, and (iii) taut cable with self-wrap contact with the object, where the application point of the wrench changes, as detailed in Fig.~\ref{fig:inst_square} for projected planar interpretation.} 
    \label{fig:thered-box-illu}
    \vspace{-1.5em}
\end{figure}

In contact-rich towing, the tether can contact the object boundary and redirect around edges (Fig.~\ref{fig:thered-box-illu}).
We refer to this as \emph{self-wrap-aware} towing: tether--object contact can become a deliberate transmission channel.
This is not merely collision avoidance: edge redirection shifts the effective wrench application point from the nominal attachment to a boundary feature, changes the moment arm, and couples routing with rigid-body motion. As a result, tension planning under a fixed straight-line tether model can be systematically biased, since both the effective length and the transmitted wrench depend on routing.
Related ``kinked tether'' effects have been modeled in tethered-UAV planning when contact is permitted \cite{xiao2018kinkedtether,cardona2021non}. For general tethered robots, routing around obstacles may require explicit topology/homotopy reasoning \cite{igarashi2010homotopic,kim2015tetheredmha,shapovalov2020tanglefree}; in contrast, this paper targets a minimal but dynamics-critical self-wrap phenomenon intrinsic to object towing, where routing-induced redirection directly changes the wrench-transmission map governing the object dynamics.


\subsection{Related Work}
Contact-implicit trajectory optimization (CITO) provides a natural abstraction for unilateral tensioning: complementarity between tension and a slackness (gap) variable lets the optimizer decide when tension becomes active without prescribing switching times \cite{posa2014contactimplicit,manchester2019variationalcito,carius2018implicitcontacts,sleiman2021dynamiccio}, and Mathematical programs with complementarity constraints (MPCC) toolchains make such formulations practical in direct transcription \cite{raghunathan2022pyrobocop}. Self-wrap-aware towing adds a second discrete structure beyond taut/slack: routing defines a transmission mode that changes both (i) the effective length map and (ii) the wrench application point/direction, hence conditioning the tensioning constraint and the rigid-body dynamics.
A strict formulation that captures routing-induced wrench transmission together with taut/slack tensioning therefore yields a mixed-integer (MI) conditioned optimal control problem (OCP) \cite{bemporadmorari1999mld,sager2009switching,sager2012integerapprox}. While principled, MI hybrid optimal control scales poorly with horizon length and is sensitive to mode scheduling \cite{sager2009switching,hogan2020reactive}, motivating smooth relaxations based on standard MPCC-to-Nonlinear Programming (NLP) regularization \cite{kanzow2013mpcc,nurkanovic2024mpccreview}. This motivates the core gap: existing tension-implicit planners typically assume fixed routing, while explicit multi-mode routing decisions can dominate the optimization landscape.

\subsection{Contributions}
Our contributions are:
(1) an OCP formulation for routing-conditioned, tensioning-implicit planning in planar cable-towed object manipulation, coupling mode-wise effective-length and wrench maps with an taut/slack tension constraint;
(2) a relaxation hierarchy from a strict mode-conditioned OCP to Full-Mode Relaxation (FMR), Binary-Mode Relaxation (BMR), and Implicit-Mode Relaxation (IMR), that trades mode explicitness for tractability and robustness;
(3) a systematic study covering in-OCP solutions and two transfer tests (open-loop rollouts on a mismatched numerical plant and MuJoCo model), showing how routing representations shape solve feasibility and self-wrap emergence (especially with IMR), and evaluating robustness on these transfer tests.




\section{Problem Statement}
We study planar cable-towed object manipulation, in which a mobile robot (modeled as a point gripper) pulls a rigid prism-shaped object via a cable anchored to one face of the object (Fig.~\ref{fig:thered-box-illu}). The cable is a unilateral transmission element and is assumed inextensible up to its rest length $L_0$: it can pull only when taut and becomes force-free when slack. During towing, the cable may contact the object boundary and redirect around a boundary \emph{edge} (a \emph{vertex} in the 2D top-view model (Fig.~\ref{fig:inst_square})), which changes the effective cable length and the wrench transmission channel. The object and gripper evolve under Newton--Euler dynamics driven by the transmitted cable wrench and ground friction; the instantiated dynamics, used in this paper, are given in App.~\ref{app:wrench}. We assume a known convex polygon cross-section, frictionless redirection at the contacted edge (tension magnitude conserved), and Coulomb ground friction with given coefficient $\mu$ in the benchmark.

We formulate a tensioning-implicit trajectory optimization (TITO) problem over a horizon $k=0,\dots,N$ that plans the robot motion while the object motion follows the transmitted cable wrench. Tensioning is enforced implicitly through a taut/slack constraint using a slackness variable $g_k:=L_0-d_k \ge 0$ with respect to the effective cable length $d_k$ and a nonnegative tension $T_k$, while routing is represented by transmission modes that define mode-wise effective-length and wrench maps. A strict mode-conditioned reference yields an MI problem; we therefore build a relaxation hierarchy (FMR/BMR/IMR) that progressively reduces explicit routing decisions while preserving the tensioning-implicit structure. 

\section{Routing-Conditioned Tensioning-Implicit Trajectory Optimization}
\label{sec:methods}
\begin{figure}[t]
    \vspace{1em}
    \centering
    \includegraphics[width=0.85\linewidth]{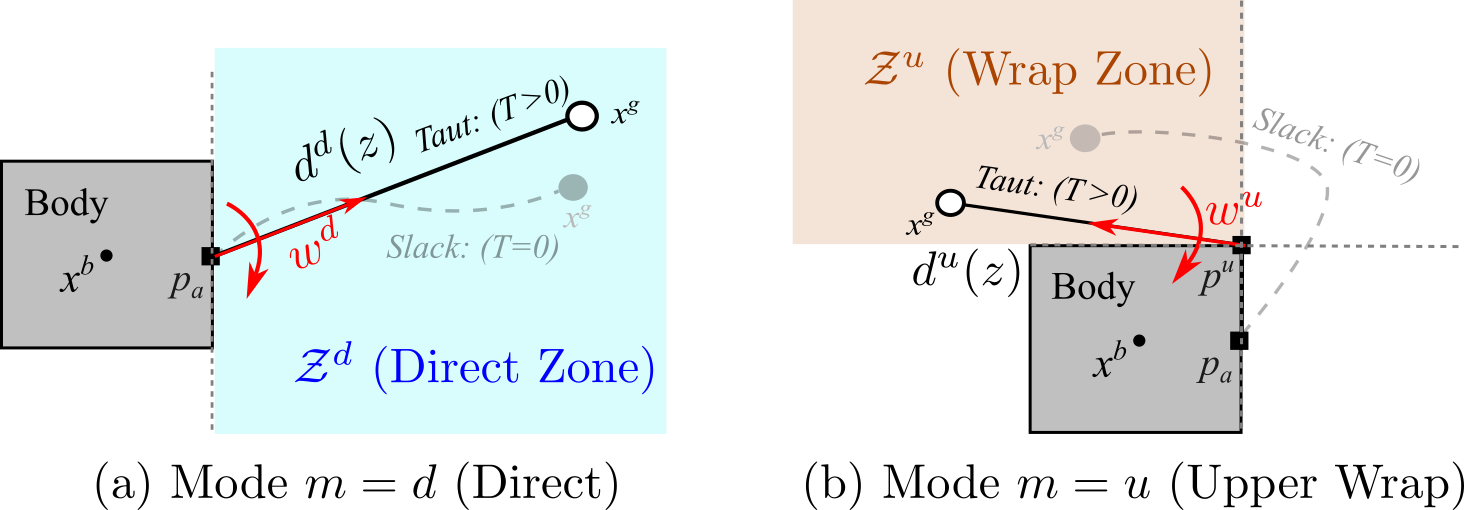}
    \caption{Illustration of cable-towed manipulation of a box in 2D (top view) at two modes: (a) direct routing mode, and (b) self-wrap routing mode.}
    \label{fig:inst_square}
    \vspace{-1em}
\end{figure}

This section first gives a general formulation of the proposed routing-conditioned TITO problem, then instantiates it for a planar square-box benchmark (Fig.~\ref{fig:inst_square}), and then presents a relaxation hierarchy from a strict mode-conditioned reference to FMR/BMR/IMR.


\subsection{General TITO problem description}
\label{sec:methods_abs}

We consider a discrete-time horizon $k=0,\dots,N$ with time step $\Delta t$. Let $z_k\in\mathbb{R}^{n_z}$ be the state of the object and robot, $u_k\in\mathbb{R}^{n_u}$ be the control input, and $T_k\in[0,T_{\max}]$ be the cable tension magnitude.
The coupled system evolves under the control input and the tension effects, which induces an object wrench $w_k\in\mathbb{R}^{n_w}$ and a robot-side cable force $\xi_k\in\mathbb{R}^{n_{\xi}}$ as
\begin{align}
z_{k+1} = f_\Delta(z_k,u_k,w_k,\xi_k),\qquad k=0,\dots,N-1.
\label{eq:abs_dyn}
\end{align}
The tension effects are closed by routing and tension, i.e., $(w_k,\xi_k)=(w(\gamma_k,z_k,T_k),\,\xi(\gamma_k,z_k,T_k))$ as defined in the following.

\subsubsection{Transmission modes and conditioned maps}
Routing is represented by a finite set of transmission modes $\mathcal{M}$ with cardinality $|\mathcal{M}|=K$. Each mode $m\in\mathcal{M}$ defines an effective-length map $d^m:\mathbb{R}^{n_z}\!\to\mathbb{R}$, transmission maps $w^m:\mathbb{R}^{n_z}\!\times\!\mathbb{R}_{\ge 0}\!\to\mathbb{R}^{n_w}$ and $\xi^m:\mathbb{R}^{n_z}\!\times\!\mathbb{R}_{\ge 0}\!\to\mathbb{R}^{n_\xi}$, and a feasibility function $\psi^m:\mathbb{R}^{n_z}\!\to\mathbb{R}^{n_{\psi}^m}$ defining $\mathcal{Z}^m:=\{z\mid \psi^m(z)\le 0\}$ componentwisely, where $\psi^m(z_k)\le 0$ are mode feasibility constraints instantiated in App.~\ref{app:feasibility} regarding the example in Sec.~\ref{sec:methods_inst}.
\begin{align}\label{eq:abs_select}
d_k &:= \sum_{m\in\mathcal{M}} \gamma_k^m\, d^m(z_k), \\
w_k &:= \sum_{m\in\mathcal{M}} \gamma_k^m\, w^m(z_k,T_k), 
\xi_k := \sum_{m\in\mathcal{M}} \gamma_k^m\, \xi^m(z_k,T_k).\nonumber
\end{align}
where $\gamma_k\in\mathbb{R}^K$ are mode weights with elements $\gamma_k^m \in \{0,1\}$ as in the strict reference; in the relaxations, $\gamma_k$ is either optimized as a continuous variable or specified implicitly (Sec.~\ref{sec:methods_ladder}).


\subsubsection{Tensioning as an implicit taut/slack law}
To present tension functioning in the system, a strict taut/slack transmission law can then be written in the standard force--gap complementarity form,
\begin{align}\label{eq:abs_comp}
T_k \ge 0,\qquad g_k \ge 0,\qquad T_k\,g_k = 0,
\end{align}
where
\begin{align}\label{eq:abs_gap}
g_k := L_0 - d_k
\end{align}
defines the effective slackness (\emph{signed distance} in CITO) using mode-aggregated effective length $d_k$ in \eqref{eq:abs_select}.
Thus, \eqref{eq:abs_comp} encodes the tensioning's complementary slack/taut constraint.

\subsubsection{Objective}
A general OCP's objective of the form
\begin{align}\label{eq:abs_cost}
J := \sum_{k=0}^{N-1}\ell_k(z_{k+1},u_k,T_k) + \ell_N(z_N),
\end{align}
where $\ell_k$ penalizes task tracking error and control effort with optional smooth regularization cost, while $\ell_N$ enforces terminal accuracy. \eqref{eq:abs_cost} is augmented by conditioning the aforementioned constraints and completed to a reference OCP in Sec.~\ref{sec:ref_ocp}.

\subsection{Instantiation: planar square-box towing benchmark}
\label{sec:methods_inst}

We instantiate the abstract interface on a planar square-box benchmark (Fig.~\ref{fig:inst_square}).
All discrete-time dynamics and mode-conditioned transmission terms $(w^m,\xi^m)$ follow App.~\ref{app:wrench};
here we only specify the geometric primitives and the resulting mode-wise effective-length maps $d^m$ that define the tensioning gap.

\subsubsection{State and geometric primitives}
The box pose around its center of mass (CoM) is $x_k^b:=(p_k^b,\theta_k)\in SE(2)$ and the gripper position is $x_k^g\in\mathbb{R}^2$.
We use the stacked state $z_k=(x_k^b,v_k^b,x_k^g,v_k^g)$.
A tether anchor point is fixed on one face at $p_a^{\mathrm{body}}\in\mathbb{R}^2$ in the body frame.
Let $\mathcal{V}:=\{\mathrm{u},\mathrm{l}\}$ denote the two adjacent vertices with body-frame coordinates
$c_{\mathrm{u}}^{\mathrm{body}},c_{\mathrm{l}}^{\mathrm{body}}\in\mathbb{R}^2$.
With planar rotation $R(\theta_k)\in SO(2)$, their world-frame positions are
\begin{align}\label{eq:inst_points}
p_{a,k} &:= p^b_k + R(\theta_k)p_a^{\mathrm{body}},\nonumber\\
p^{v}_{k} &:= p^b_k + R(\theta_k)c_{v}^{\mathrm{body}},\qquad v\in\mathcal{V}.
\end{align}

\subsubsection{Mode-wise effective-length maps}
We use three routing modes $\mathcal{M}=\{\mathrm{d},\mathrm{u},\mathrm{l}\}$: direct ($\mathrm{d}$) and single-vertex redirection via $v\in\mathcal{V}$.
The mode-wise effective-length maps are
\begin{align}\label{eq:inst_lengths}
d^{\mathrm{d}}(z_k) &:= \|x^g_k-p_{a,k}\|_\varepsilon,\nonumber\\
d^{v}(z_k) &:= \|p^{v}_{k}-p_{a,k}\|_\varepsilon + \|x^g_k-p^{v}_{k}\|_\varepsilon,\qquad v\in\mathcal{V}.
\end{align}
Where $\|r\|_\varepsilon:=\sqrt{r^\top r+\varepsilon}$ and $\hat r_\varepsilon:=r/\|r\|_\varepsilon$ with a small $\varepsilon>0$ has been used in this work. The aggregated effective length $d_k$ and slackness $g_k$ follow from \eqref{eq:abs_select}--\eqref{eq:abs_gap}.
Mode-conditioned transmission and feasibility constraints are given in App.~\ref{app:wrench}--\ref{app:feasibility}.
\subsection{Relaxation hierarchy softened from reference OCP}
\label{sec:methods_ladder}

We start from a strict mode-conditioned reference in which routing is discrete and tensioning follows \eqref{eq:abs_comp}, yielding an MI formulation. To obtain tractable smooth NLPs, we build a relaxation hierarchy that progressively relaxes (i) the mode representation in \eqref{eq:abs_select} and (ii) the enforcement of \eqref{eq:abs_comp}.
All variants share the same dynamics template \eqref{eq:abs_dyn} and the base task objective $J$ in \eqref{eq:abs_cost}. Each relaxation augments $J$ with additional
penalties/slacks whose weights are tightened stagewise.

\subsubsection{Reference Optimal Control Problem (OCP)}
\label{sec:ref_ocp}
We optimize over the state trajectory $z_{0:N}$, control $u_{0:N-1}$, tension $T_{0:N-1}$, one-hot mode indicators $\delta_{0:N-1}$, and linear complementarity slack penalty $\varepsilon_{0:N-1}$ by means of the proposed MI--MPCC-like OCP formed in
\begin{subequations}\label{eq:ref_ocp}
\begin{align}
\min_{\{z,u,T,\delta,\varepsilon\}}\quad & \mathcal{J}
\label{eq:ref_ocp_obj}\\
\text{s.t.}\quad
& z_{k+1}=f_\Delta(z_k,u_k,w_k),
\label{eq:ref_ocp_dyn}\\
& \delta_k\in\{0,1\}^{K},\ \mathbf{1}^\top\delta_k=1,
\label{eq:ref_ocp_onehot}\\
& (d_k,w_k)\ \text{given by \eqref{eq:abs_select} with }\gamma_k=\delta_k,
\label{eq:ref_ocp_closure}\\
& T_k\ge 0,\ g_k\ge 0,\label{eq:ref_ocp_T_g_nonegative} \\
& T_k g_k\le \varepsilon_k,\ \varepsilon_k\ge 0,
\label{eq:ref_ocp_slackcomp}\\
& \psi^m(z_k)\le M_{\psi}(1-\delta_k^m),
\label{eq:ref_ocp_feas}\\
& d^m(z_k)-L_0\le M_{\ell}(1-\delta_k^m),
\label{eq:ref_ocp_inext}\\
& z_0=\bar z_0,\ z_k\in\mathcal{X},\ u_k\in\mathcal{U},\ 0\le T_k\le T_{\max}.
\label{eq:ref_ocp_bounds}
\end{align}
\end{subequations}
where $\gamma_k=\delta_k$ is conditioned to form the tether wrench $w_k$ as closed by \eqref{eq:abs_select}.
The slack variable $\varepsilon_k$ relaxes taut/slack tensioning and is penalized via $J_{\varepsilon}:=\sum_{k=0}^{N-1}\varepsilon_k$ with weight $\lambda_{\varepsilon}>0$, i.e., $\mathcal{J}:=J+\lambda_{\varepsilon}J_{\varepsilon}$+(\text{task-specific regularizers}).
The big-$M$ constraints \eqref{eq:ref_ocp_feas}--\eqref{eq:ref_ocp_inext} activate mode-wise feasibility and inextensibility
only for the selected mode.

\subsubsection{Relaxation A: Full-Mode Relaxation (FMR)}
\label{sec:FMR}

FMR removes integer variables by relaxing the one-hot constraint \eqref{eq:ref_ocp_onehot} to the simplex.
The decision variables are $\{z_{0:N},u_{0:N-1},T_{0:N-1},\delta_{0:N-1},\varepsilon_{0:N-1}\}$,
with $\delta_k$ constrained to the simplex
\begin{align}
\delta_k \in \Delta^K := \{\delta\in\mathbb{R}^K\mid \delta\ge 0,\ \mathbf{1}^\top\delta=1\}.
\label{eq:FMR_simplex}
\end{align}
All other constraints of the reference OCP \eqref{eq:ref_ocp} are kept, while the cost is modified as $\mathcal{J}:=J+\lambda_{\varepsilon}J_{\varepsilon}
+\lambda_{\mathrm{1hot}}J_{\mathrm{1hot}}$,
where $J_{\varepsilon}$ is inherited from \eqref{eq:ref_ocp_obj} and $\lambda_{\mathrm{1hot}}>0$, that augments to suppress mixed-mode artifacts and expressed as
\begin{align}\label{eq:FMR_j_onehot}
J_{\mathrm{1hot}} &:= \sum_{k=0}^{N-1}\sum_{m\in\mathcal{M}} \delta_k^m(1-\delta_k^m).
\end{align}

\subsubsection{Relaxation B: Binary-Mode Relaxation (BMR)}
\label{sec:BMR}
BMR collapses the $K$-ary mode choice to a binary \emph{direct/redirected} decision.
The decision variables are $\{z_{0:N},u_{0:N-1},T_{0:N-1},\delta^{\mathrm{r}}_{0:N-1}\}$, where
\begin{align}
\delta_k^{\mathrm{r}}\in\{0,1\},\qquad \delta_k^{\mathrm{d}}:=1-\delta_k^{\mathrm{r}},\qquad
\gamma_k := (\delta_k^{\mathrm{d}},\delta_k^{\mathrm{r}}).
\label{eq:BMR_binary}
\end{align}
The routing closure \eqref{eq:abs_select} is applied over the reduced mode set $\mathcal{M}_{2}:=\{\mathrm{d},\mathrm{r}\}$
using $\gamma_k$ in \eqref{eq:BMR_binary}. A smooth vertex selector to instantiate $(d^{\mathrm r},w^{\mathrm r})$ in the square-box case is elaborated in App.~\ref{app:feasibility}. To obtain smooth gradient with further smoothness, we substitute \eqref{eq:ref_ocp_slackcomp} by applying 
\begin{equation}\label{eq:BMR_j_eta}
    J_{\eta} := \sum_{k=0}^{N-1}\sum_{m\in\mathcal{M}_2} \delta_k^m\,\eta^2\!\big(T_k,g_k^{m}\big)
\end{equation}
with nonlinear complementarity residual (NCR) $\eta(T,g):=\sqrt{T^2+g^2+\varepsilon_{\eta}^2}-(T+g)$ with a complementarity allowance $\varepsilon_{\eta}=1e-6$, which eliminates the previous linear complementarity slack penalty in the decision set.

Relative to \eqref{eq:ref_ocp}, BMR replaces the $K$-ary feasibility and inextensibility indicators
\eqref{eq:ref_ocp_feas}--\eqref{eq:ref_ocp_inext} by the same big-$M$ form over $\mathcal{M}_2$ with modification (App.~\ref{app:feasibility}). Eventually, $\mathcal{J}:=J+ \lambda_{\eta}J_{\eta}$ is used\footnote{\(J_{\mathrm{1hot}}\) evaluated on \(\mathcal{M}_2\) is imposed to 0 by definition.}, with $\lambda_{\eta}>0$.

\subsubsection{Relaxation C: Implicit-Mode Relaxation (IMR)}
\label{sec:IMR}
IMR removes routing modes from the decision set with
new decision variables $\{z_{0:N},u_{0:N-1},T_{0:N-1}\}$.
Instead of optimizing $\gamma_k$, IMR represents a smooth routing gate $\sigma_k=\sigma(z_k)\in(0,1)$ and sets
\begin{align}
\gamma_k := (\sigma_k,1-\sigma_k)\qquad \text{over }\mathcal{M}_2:=\{\mathrm{d},\mathrm{r}\}.
\label{eq:IMR_gamma}
\end{align}
The closure \eqref{eq:abs_select} is evaluated with \eqref{eq:IMR_gamma}. As in BMR, redirected maps for IMR are instantiated from the same vertex selector. Moreover, mode feasibility and inextensibility are enforced via continuously gated big-$M$ constraints obtained from
\eqref{eq:ref_ocp_feas}--\eqref{eq:ref_ocp_inext} by replacing $(\delta_k^{\mathrm{d}},\delta_k^{\mathrm{r}})$ with
$(\sigma_k,1-\sigma_k)$ as detailed in App.~\ref{app:feasibility}. We inherit the NCR from BMR and rephrase our cost to
$\mathcal{J}:=J+\lambda_{\eta}J_{\eta}$.
A mode-consistent gap construction aligned with the gated inequalities is summarized in App.~\ref{app:ci_gaps}.

\begin{figure*}[t]
    \centering
    \includegraphics[width=0.75\linewidth]{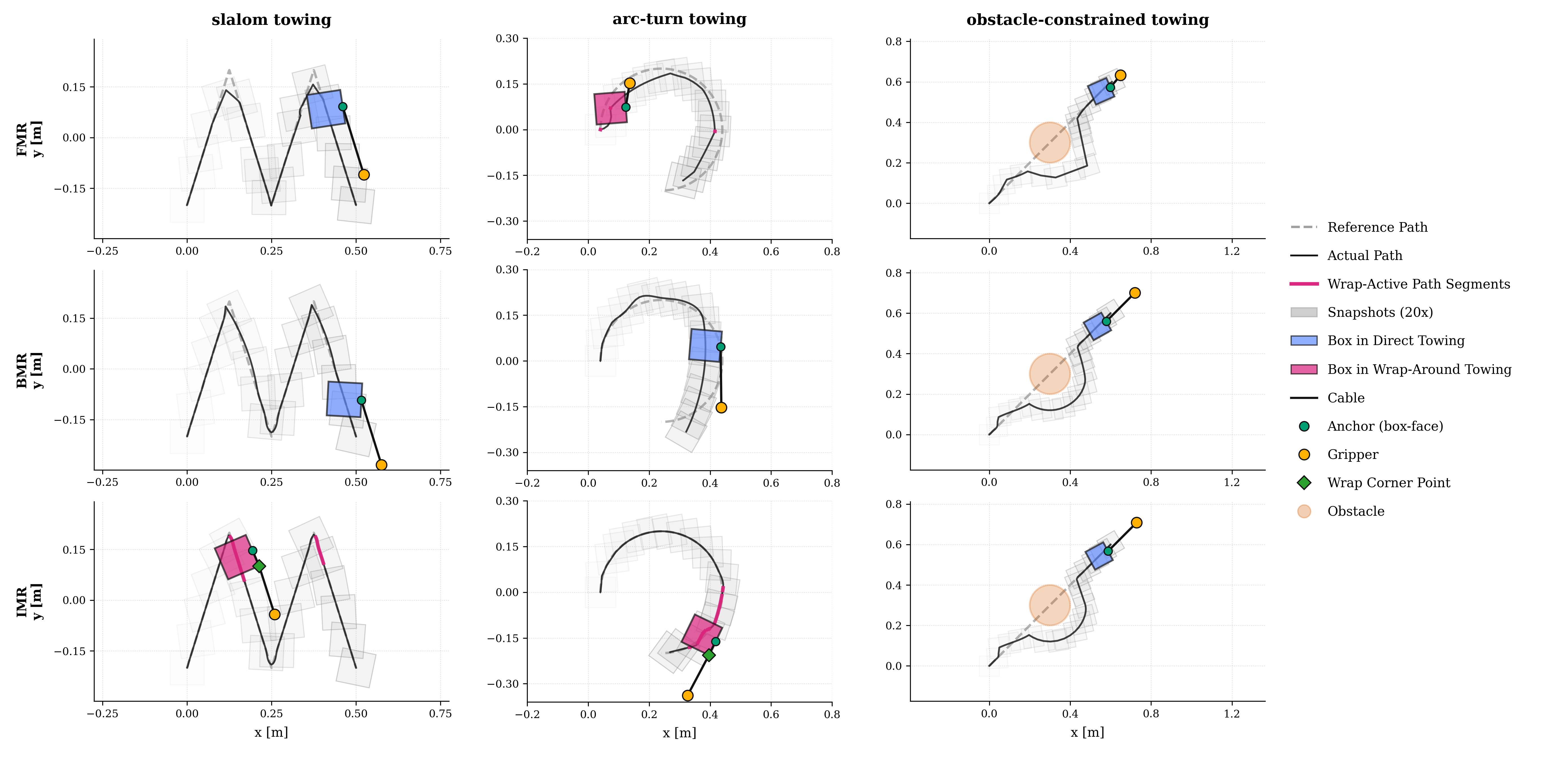}
    \caption{Representative planar towing rollouts for three scenes (columns) and three formulations (rows: FMR, BMR, IMR). Dashed: reference box CoM path. Solid: actual box CoM path with wrap-active segments highlighted. Snapshots show box poses; the cable and the active redirection vertex are overlaid when applicable. The results are shown with the same initial condition (e.g. \(x^g_{k=0}, T_{k=0}\), etc.) and model parameters.
    }
    \label{fig:spatial-xy-traj}
\end{figure*}

\begin{figure*}
    \centering
    \includegraphics[width=0.85\linewidth]{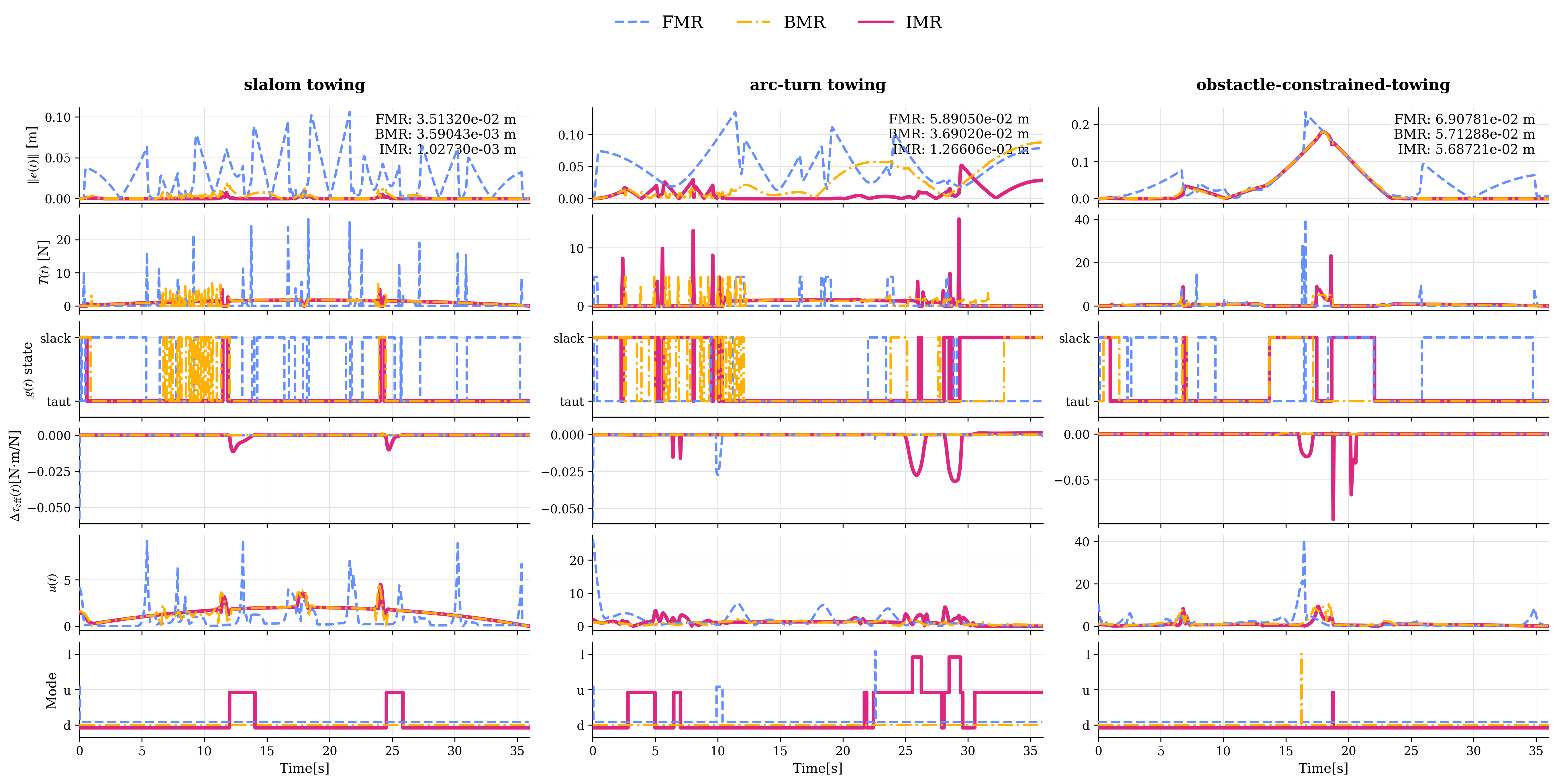}
    \caption{Time histories for the rollouts in Fig.~\ref{fig:spatial-xy-traj} (columns). From top: tracking error $\|e(x)\|$, tension $T$, taut/slack state (from gap $g$; cf.\ \eqref{eq:abs_gap}, \eqref{eq:ref_ocp_slackcomp}), effective torque-channel difference $\Delta\tau_{\mathrm{eff}}$, control action $u$, and routing mode.}
    \label{fig:spatial-state-traj}
\end{figure*}

\begin{table*}[t]
\centering
\caption{\textbf{Randomized-Initialization Benchmark (N=15 per scene-method).}
Success policy: slalom towing uses strict \texttt{task\_success} reported by solver; Arc-turn uses
$\mathrm{RMSE}\le 0.08,\ \mathrm{FinalErr}\le 0.10,\ \mathrm{WrapRatio}\ge 0.05$;
Constrained-obstacle towing uses $\mathrm{RMSE}\le 0.10,\ \mathrm{FinalErr}\le 0.12$.
Notations: SR (Success Rate), Time (solver time), RMSE (tracking error), Wrap (wrap ratio).}
\label{tab:benchmark_3methods}
\setlength{\tabcolsep}{4pt}
\renewcommand{\arraystretch}{1.3}
\resizebox{0.95\textwidth}{!}{%
\begin{tabular}{l|cccc|cccc|cccc}
\hline
\hline
\textbf{Scene} &
\multicolumn{4}{c|}{\textbf{FMR}} &
\multicolumn{4}{c|}{\textbf{BMR}} &
\multicolumn{4}{c}{\textbf{IMR}} \\
\cline{2-13}
& SR (\%) & Time (s) & RMSE (m) & Wrap (\%)
& SR (\%) & Time (s) & RMSE (m) & Wrap (\%)
& SR (\%) & Time (s) & RMSE (m) & Wrap (\%) \\
\hline
\textbf{Zigzag}
& 0.0   & $84.08 \pm 1.01$   & --       & --
& 100.0 & $64.97 \pm 0.35$   & $0.0037$ & 0.0
& 86.7  & $35.33 \pm 13.34$  & $0.0017$ & 9.1 \\
\textbf{Arc Wrap}
& 0.0   & $92.36 \pm 5.06$   & --       & --
& 100.0 & $83.36 \pm 4.05$   & $0.0388$ & 0.7
& 73.3  & $80.12 \pm 8.04$   & $0.0299$ & 16.0 \\
\textbf{Obstacle}
& 0.0   & $107.67 \pm 5.90$  & --       & --
& 100.0 & $81.29 \pm 4.78$   & $0.0571$ & 0.3
& 80.0  & $158.07 \pm 51.96$ & $0.0570$ & 1.0 \\
\hline
\hline
\end{tabular}%
}
\end{table*}

\begin{figure*}
    \centering
    \includegraphics[width=0.85\linewidth]{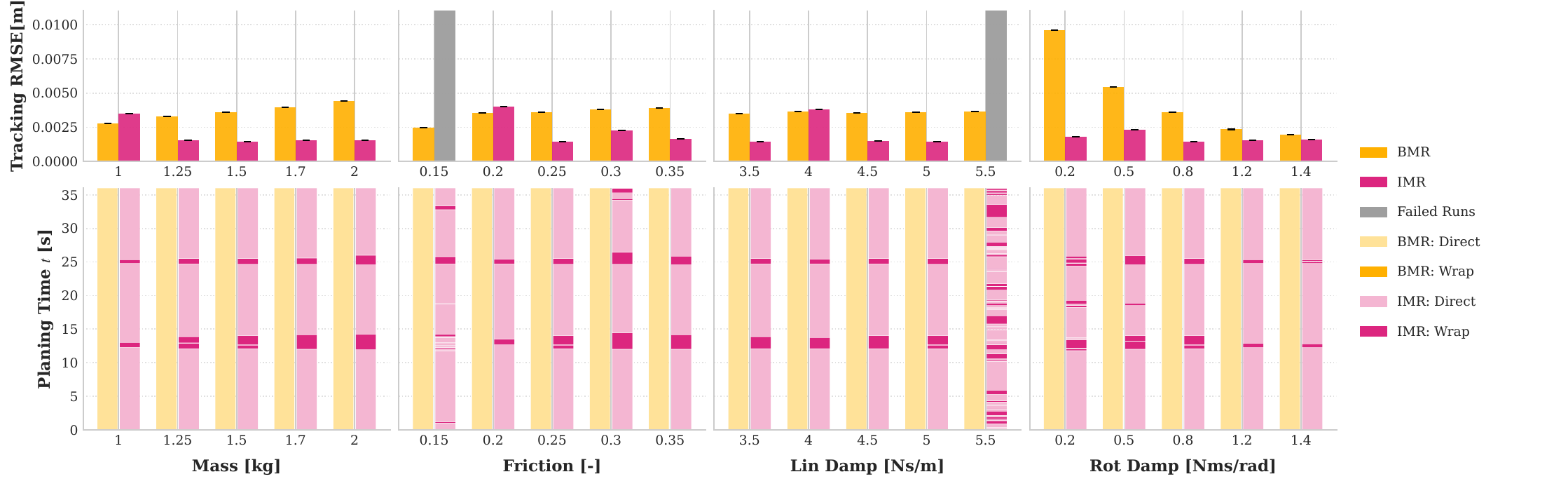}
    \caption{Parameter sweep on the slalom benchmark for BMR and IMR. Top: tracking RMSE (gray indicates failed runs). Bottom: planning of direct/redirected (wrap-active) mode.}
    \label{fig:parameter-ocp-benchmarking}
\end{figure*}

\begin{figure}
    \centering
    \includegraphics[width=1\linewidth]{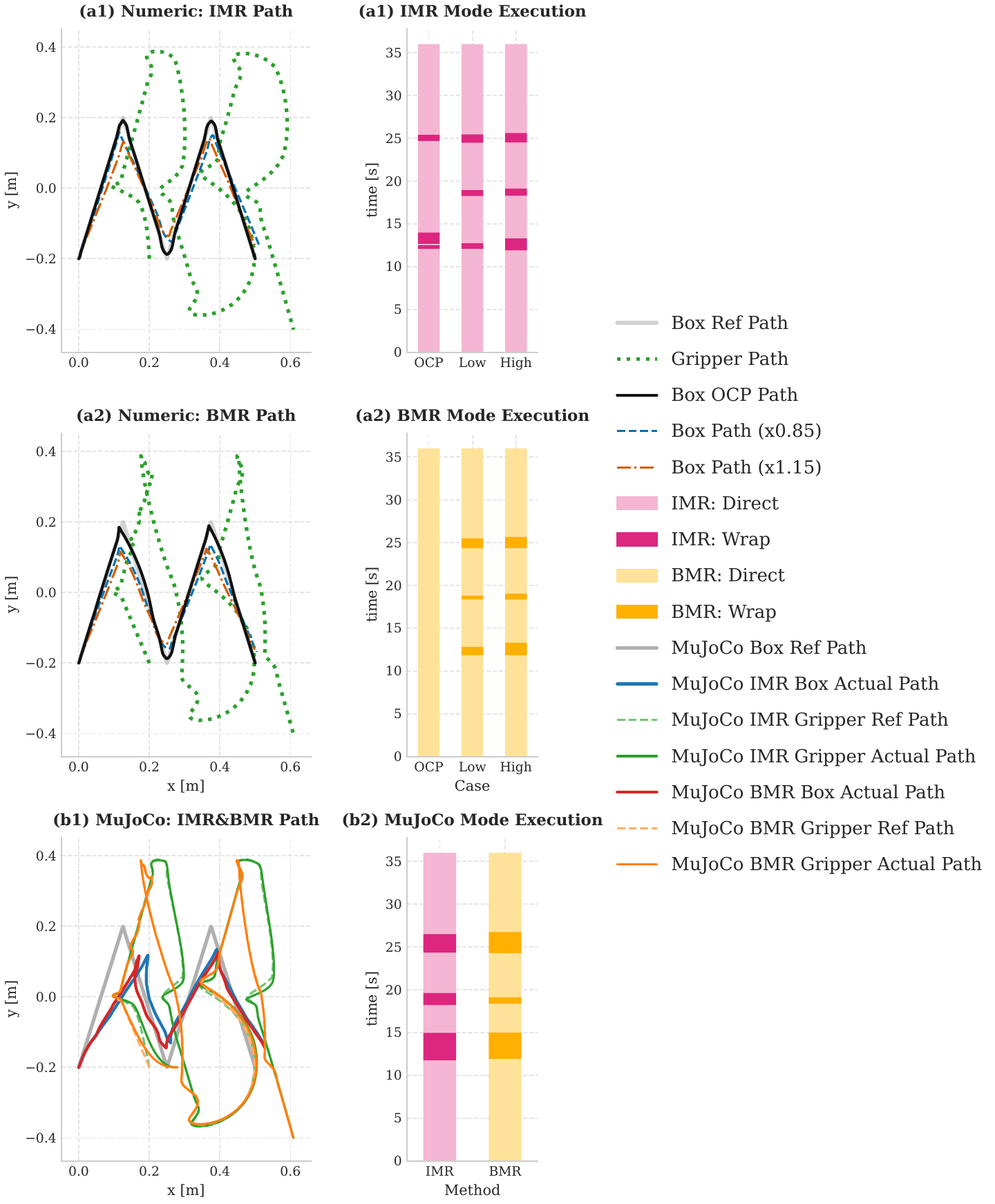}
    \caption{Open loop trajectory validation on ``real" numeric plant for BMR and IMR with mismatched plant parameters (0.85 and 1.15 fraction of the baseline parameters) and \textsc{Mujoco} simulation. Results focus on spatial trajectory and recorded wrap behavior. 
    }
    \label{fig:real-numerial-plant}
    \vspace{-1em}
\end{figure}

\section{Implementation and Simulation Results}
\label{sec:results}

The proposed relaxation hierarchy is evaluated on three planar towing tasks that progressively stress routing-conditioned wrench transmission: (i) \emph{slalom towing} (zigzag), (ii) \emph{arc-turn towing} (circular), and (iii) \emph{obstacle-constrained towing} (straight-path with a hard obstacle-avoidance constraint). Each task is solved using three formulations: FMR, BMR, and IMR (Sec.~\ref{sec:methods_ladder}). Representative trajectories and time histories are shown in Figs.~\ref{fig:spatial-xy-traj}--\ref{fig:spatial-state-traj}; randomized-initialization benchmarking is summarized in Table~\ref{tab:benchmark_3methods}; robustness/transfer are reported in Figs.~\ref{fig:parameter-ocp-benchmarking}--\ref{fig:real-numerial-plant}.

All tasks share the same dynamics discretization, horizon, and base task objective $J$ (Sec.~\ref{sec:methods_abs}--\ref{sec:methods_ladder}). We use multi-shooting with $N=600$ and $\Delta t=60$,ms (36,s), a semi-implicit integration scheme, and solve the resulting NLPs in CasADi \cite{Andersson2019} with IPOPT. Unless otherwise stated, common solver options and objective weights are kept identical across FMR, BMR, and IMR; only the relaxation-specific terms in Sec.~\ref{sec:methods_ladder} differ. All runs are executed on an Intel i5-13400F workstation, and we report wall-clock solver time.

\subsection{Mode execution and self-wrap needs}
Fig.~\ref{fig:spatial-xy-traj} summarizes qualitative behaviors across the three scenes, and Fig.~\ref{fig:spatial-state-traj} shows the corresponding time histories of state, tensioning, and mode-related diagnostics.

FMR frequently settles on mixed and short-lived wrap segments, which leads to visibly distorted cable paths and large tracking errors in both slalom and arc-turn. This is accompanied by unstable time histories, including chattering in tension/tautness and aggressive control actions (Fig.~\ref{fig:spatial-state-traj}). In contrast, the two formulations with reduced mode representation produce more consistent behaviors. BMR typically completes the tasks with a conservative strategy that stays mostly in the direct mode and close to the direct/redirect boundary, as indicated by stable mode and tautness traces across scenes. IMR, on the other hand, tends to maintain redirected (wrap-active) segments when additional rotational authority is needed. This difference is visible in the sharp zigzag turns, where IMR produces more distinct wrap engagement while BMR tracks with a rounder motion. In arc-turn, IMR maintains a long wrap-active segment in the second half of the path, whereas BMR remains largely in the direct mode.

Overall, the results support the following qualitative conclusion: reducing explicit routing decision variables mitigates mixed-mode artifacts and mode-chattering, and IMR more reliably produces sustained self-wrap when redirection is structurally beneficial for turning, while maintaining physically consistent tensioning.

The torque-channel diagnostic $\Delta\tau_{\mathrm{eff}}(t)$ (App.~\ref{app:tau_diag}) further clarifies this behavior. It stays near zero when the trajectory remains in the direct mode and becomes active during wrap episodes. In IMR, the activation is clean and sustained, aligned with the wrap segments in Fig.~\ref{fig:spatial-xy-traj}, consistent with the fact that the mode-conditioned wrench application point strongly affects turning performance. Finally, an obstacle-avoidance scene is included to demonstrate feasibility under additional constraints, even though this scene does not necessarily require self-wrap.

\subsection{Benchmarking of optimization}
Feasibility and performance are evaluated under randomized initializations by perturbing the initial gripper pose and the solver initial guesses for tension and mode variables (Table~\ref{tab:benchmark_3methods}). For each scene--method pair, we run $N{=}15$ trials and report success rate (SR), solver time, tracking root mean square error (RMSE), and wrap engagement over the full horizon using the task-specific success policies in the table caption.

FMR is brittle under these perturbations, achieving $0\%$ SR across all tasks with long solve times and unstable solutions. Between the two feasible variants, BMR is more solver-reliable, while IMR achieves comparable RMSE
and produces stronger and more sustained wrap engagement when redirection is beneficial. In short, BMR tends to solve more reliably, whereas IMR more consistently activates the redirected wrench channel, which provides additional rotational leverage during turning.

To assess sensitivity beyond initialization effects, key physical parameters are swept on the slalom benchmark (chosen for clearer wrap distinction) for BMR and IMR (Fig.~\ref{fig:parameter-ocp-benchmarking}). The same trade-off persists: BMR remains more solver-reliable, while IMR is more likely to activate wrap and can slightly improve tracking when self-wrap is beneficial. The sweep also shows clear shifts in wrap timing. Increasing the mass (and thus inertia) generally prolongs wrap-active segments, consistent with a larger turning benefit from redirection. Reducing rotational damping makes the box more agile, and IMR tends to trigger additional wrap events; for example, an extra wrap at the second zigzag turn appears under low damping but not in the baseline. These behaviors are muted in BMR, which often completes the task with a mostly direct-mode strategy even when IMR elects to redirect.

\subsection{Open-loop Trajectory Validations}
\label{sec:openloop_mismatch}

Open-loop transfer is evaluated on a ``numeric plant'' that intentionally differs from the OCP model in two aspects:
(i) routing is classified strictly as direct/redirected (no blending), and
(ii) tension is generated by a unilateral spring--damper law with stiffness $k_c$ and damping $b_c$ as
\begin{equation}
T_k=\mathrm{sat}_{[0,T_{\max}]}\!\Big(k_c\big(d_{\mathrm{k}}-L_0\big)+b_c\,\dot d_{\mathrm{k}}\Big),
\end{equation}
instead of the slack-relaxed tensioning constraint used in the OCP.
The planned gripper trajectory $(x^g,v^g)$ is tracked by a PD controller on the gripper dynamics, while the box evolves under the transmitted cable wrench and frictional contact.

Fig.~\ref{fig:real-numerial-plant} reports rollouts under nominal conditions and under $\pm 15\%$ dynamical parameter scaling.
Despite the mode and tension-model mismatch, IMR typically preserves task completion and maintains a consistent routing state; the executed gripper motion remains smooth and arc-like, matching the planned sustained redirection.

For BMR, even when the optimized solution contains little planned wrap-routings, the plant can still enter self-wrap due to discrepancies in rotational dynamics.
Correspondingly, BMR plans often look more aggressive than IMR (less ``round'' and more ``shortcut''), suggesting that BMR may trade sustained wrap engagement for faster gripper repositioning to achieve tracking.

Finally, the same open-loop trajectory is replayed in \textsc{MuJoCo}.
A discretized cable is modeled as a chain of \texttt{capsule} bodies, using moderate soft-constraint settings \texttt{solref="0.020 1"} and \texttt{solimp="0.72 0.88 0.010"}, since \textsc{MuJoCo} does not provide a dedicated high-fidelity soft-contact solver for cable mechanics.
Other model parameters are kept as close as possible to the OCP model.
While the planned trajectories still induce turn-aware self-wrap qualitatively shown similarly on numeric plant, the executed motions can deviate substantially: the behavior is sensitive to \textsc{MuJoCo}'s constraint solver and to cable bending effects introduced by the discretized rope, even after soft-parameter tuning.
This gap motivates contact/tensioning models beyond purely algebraic complementarity for physics-engine transfer, and underscores the remaining challenges for real-world deployment.

\section{Discussion}
\label{sec:discussion}

\subsection{How mode preservation affects optimality}
Across the three scenes/tasks, FMR/BMR/IMR show clearly different failure modes and routing patterns (Figs.~\ref{fig:spatial-xy-traj}--\ref{fig:spatial-state-traj}, Table~\ref{tab:benchmark_3methods}) even though they share the same physical primitives. In practice, the choice of mode representation matters as much as the physics: taut/slack tensioning is already hybrid, and routing redirection changes both the effective-length map and the wrench map, so the active constraints seen by the solver can change abruptly during the solve. When routing is optimized explicitly (simplex or binary), the solver must handle these abrupt map changes while also “hesitating” to switch, because switching immediately trades tracking cost against feasibility/penalties. This naturally leads to mode-avoidance near switching boundaries and explains why the full mode-preserving FMR is brittle under randomized initializations.

Between the two workable variants, BMR is typically easier to solve, while IMR produces wrap episodes more consistently. This points to a middle ground: keep only a small set of explicit routing decisions (or use a surrogate selector) to avoid boundary-hugging local minima, without committing to a full discrete schedule. IMR removes the combinatorial burden of mode scheduling, but routing transitions are still not “free”: the gate changes the effective constraints and local conditioning through state-dependent mixing. Empirically, this makes sustained wrap segments more likely when redirection is actually useful.

The arc-turn scene also highlights a structural limitation of single-cable towing. The object wrench is generated by a nonnegative scalar tension together with redirection (App.~\ref{app:wrench}), so the achievable wrench set is low-dimensional and further restricted by routing feasibility (App.~\ref{app:feasibility}). If the task needs a sustained torque profile to stay time-aligned with the reference, this limited wrench channel may match the required profile only intermittently. As a result, the optimizer may choose short redirected segments that fix the spatial contour but do not maintain one-to-one time alignment. In some runs, the same effect appears as brief, high-acceleration gripper bursts and rapid routing toggles; this is consistent with the current objective, which does not explicitly penalize switching frequency or enforce actuator/slew limits beyond generic smoothness terms.

\subsection{Limitations and future work}
This study focuses on the simplest self-wrap effect that still changes the dynamics. Routing is limited to a few redirection choices around one anchored face in a simple geometry; general convex boundaries are not addressed. Richer routing (multiple redirections, multiple contacted edges, repeated windings) would need extra geometric primitives or an explicit mode-transition model, but naively enumerating these routing possibilities quickly makes the OCP blow up in dimension. 

The current transmission model also assumes frictionless redirection at a vertex/edge, so tension magnitude is conserved. Real cables can have edge friction, compliance, and distributed contact, which change both the effective-length map and the wrench map; capturing this may require extra state (e.g., a boundary contact coordinate) or a reduced-order continuum model. Finally, scenes with nontrivial wrap patterns usually need online replanning: an offline OCP alone is not sufficient, e.g. the robot may cause undesired contact with the object due to biased trajectory (see complementary video in \textsc{Mujoco} test). A practical next step is to warm-start MPC from offline solutions, or to design a tractable MPC problem that preserves the same tensioning/routing feasibility structure.

\section{Conclusion}
This paper formulates self-wrap-aware cable-towed manipulation as a trajectory-optimization problem in which routing changes the effective cable length and wrench transmission, and tensioning is enforced implicitly through a taut/slack constraint. A relaxation hierarchy is introduced, moving from a strict mode-conditioned reference to three tractable formulations (FMR/BMR/IMR), and evaluated systematically on planar towing benchmarks. The results indicate that, in turning-dominant scenes, purposeful self-wrap contact can be required to achieve good tracking under single-cable towing, but that the chosen mode representation strongly shapes solver behavior (e.g., mixed-mode artifacts, mode avoidance, and wrap consistency). The study also highlights practical directions to improve robustness and deployability, including alternative mode representations with weaker mode-conditioning and online replanning via MPC.

\appendix

\subsection{Instance dynamics and Mode conditioned wrench transmission}
\label{app:wrench}
The composed object and robot dynamics instantiated in this work are described by 
\begin{subequations}\label{eq:inst_dyn_disc_short}
\begin{align}
v^b_{k+1} &= v^b_k + \Delta t\, I^{-1}\!\Big(w(z_k,T_k)-\zeta_{ext}(v^b_k)\Big), \label{eq:obj_dyn_v_short}\\
v^g_{k+1} &= v^g_k + \Delta t\, m^{-1}\!\Big(u_k-b v^g_k-\xi(z_k,T_k)\Big). \label{eq:rob_dyn_v_short}
\end{align}
\end{subequations}
where \(I \in \mathbb{R}^{3\times3},m\in \mathbb{R}^{2\times2}\)  are the object and gripper inertia matrix, respectively. \(\zeta_{ext}(\dot{x}^b_k) \in \mathbb{R}^{3\times1}\) presents the external wrench incl. damping and dry friction (represented by Coulomb model), \(w(z_k,T_k)\) is the applied wrench conditioned by wrap indexed by $w^m(z_k,T_k)$ following. \(b\) is the damping factor of the gripper dynamics, and \(\xi^m(z_k,T_k)\) presents the tension effect on gripper.

The mode conditioned wrench $w^m(z_k,T_k)$ used in
Sec.~\ref{sec:methods_inst} is detailed as:
\emph{Direct mode ($m=\mathrm{d}$).}
Let $r_k^{ag}:=x_k^g-p_{a,k}$ and $e_k^{ag}:=\widehat{r_k^{ag}}_\varepsilon$.
Let $r_k^{a}:=p_{a,k}-x_k^b$ be the anchor moment arm about $x_k^b$. $w^d(z_k,T_k)$ contains
\begin{align}
F_{b,k}^{\mathrm{d}} &= T_k\,e_k^{ag}, &
\tau_{b,k}^{\mathrm{d}} &= r_k^{a}\times F_{b,k}^{\mathrm{d}}.
\label{eq:wrench_direct_app}
\end{align}
While \(\xi^d(z_k,T_k) = -F_{b,k}^{\mathrm{d}}\) is used.
\emph{Wrap modes ($m\in\mathcal{V}=\{\mathrm{u},\mathrm{l}\}$).}
Fix a vertex label $v\in\mathcal{V}$ and define
$r_k^{av}:=p_k^{v}-p_{a,k}$, $r_k^{vg}:=x_k^{g}-p_k^{v}$, and
$e_k^{av}:=\widehat{r_k^{av}}_\varepsilon$, $e_k^{vg}:=\widehat{r_k^{vg}}_\varepsilon$.
Let $r_k^{v}:=p_k^{v}-x_k^b$ be the vertex moment arm.
$w^v(z_k,T_k)$ composes of
\begin{align}
F_{b,k}^{v} &= F_{b,k}^{a,v}+F_{b,k}^{v,v}, &
\tau_{b,k}^{v} &= r_k^{a}\times F_{b,k}^{a,v}+r_k^{v}\times F_{b,k}^{v,v}, \nonumber
\end{align}
where \(F_{b,k}^{a,v} = T_k\,e_k^{av},
F_{b,k}^{v,v} = T_k(-e_k^{av}+e_k^{vg})\) are used. While \(\xi^v(z_k,T_k) = - (F_{b,k}^{a,v} + F_{b,k}^{v,v})\) is specified.

For the case of BMR/IMR, we evaluate a redirected map $(d^{\mathrm{r}},w^{\mathrm{r}})$ by replacing the discrete vertex point $p_k^{v}$
above with a state-induced redirection point $p_{\rho,k}$ on the anchor-face boundary
(App.~\ref{app:feasibility}).

\subsection{Instance of wrap mode feasibility}
\label{app:feasibility}

This appendix instantiates the feasibility functions $\psi^m$ used in the big-$M$ indicators
\eqref{eq:ref_ocp_feas} and the benchmark wedge constraints (Fig.~\ref{fig:inst_square}).
We present the right-face anchor case for an instance.

\emph{Anchor-attached frame.}
Let $n_{x,k}:=R(\theta_k)[1,0]^\top$ be the outward normal of the anchor face
and $n_{y,k}:=R(\theta_k)[0,1]^\top$ be the tangential direction in world frame.
\emph{Wrap feasibility in explicit mode.}
The wrap feasibility set is instantiated by partitioning geometrical observations within the simple planar case by:
\begin{align}\label{eq:wedge_app}
    \mathcal{Z}^d: &\quad n_{x,k}^{\top}\widehat{r_k^{ag}}_\varepsilon\ge 0. \\
    \mathcal{Z}^u:&\quad n_{x,k}^{\top}\widehat{r_k^{vg}}_\varepsilon\le 0,\  \ -n_{y,k}^{\top}\widehat{r_k^{vg}}_\varepsilon\le 0 \nonumber \\
    \mathcal{Z}^l:&\quad n_{x,k}^{\top}\widehat{r_k^{vg}}_\varepsilon\le 0,\ \ n_{y,k}^{\top}\widehat{r_k^{vg}}_\varepsilon\le 0 \nonumber
\end{align}
This also implies a non-penetration constraint with respect to planar object case. In BMR, only \(\mathcal{Z}^d\) is preserved and modified as binary partition
\begin{align}
n_{x,k}^{\top} \widehat{r_k^{ag}}_\varepsilon \ge -M_{\mathrm{hs}}\big(1-\delta_k^{\mathrm d}\big),
n_{x,k}^{\top} \widehat{r_k^{ag}}_\varepsilon \le  M_{\mathrm{hs}}\big(1-\delta_k^{\mathrm r}\big),
\label{eq:bmr_partition_bigM}
\end{align}
$M_{\mathrm{hs}}>0$ is a sufficiently large constant. While, to complete the wrap feasibility, we evaluate 
\begin{align}\label{eq:bmr_wedge_bigM}
n_{x,k}^{\top} \widehat{r_k^{\rho g}}_\varepsilon \ge -M_{\mathrm{w}}\big(1-\delta_k^{\mathrm r}\big),
\bar n_{y,k}^{\top} \widehat{r_k^{\rho g}}_\varepsilon \ge -M_{\mathrm{w}}\big(1-\delta_k^{\mathrm r}\big),
\end{align}
with sufficient large $M_{\mathrm{w}}$. This construction of the constraints is followed by introducing a wrap-vertices selector. 
\begin{align}\label{eq:rho_def_app}
\rho_k := \tfrac12\big(1+\tanh(k_{\mathrm{sel}}\,\widehat{r^{ag}_k}_\varepsilon)\big) \in [0,1]
\end{align}
thus, the wrapped vertex is interpolated by
\begin{align}\label{eq:prho_def_app}
p_{\rho,k}:=x_k^b+R(\theta_k)\big(\rho_k c_{\mathrm{u}}^{\mathrm{body}}+(1-\rho_k)c_{\mathrm{l}}^{\mathrm{body}}\big).
\end{align}
\eqref{eq:bmr_wedge_bigM} is completed by selecting \(\bar n_{y,k}\) correspondent to \(p_{\rho,k}\) localization and \(r_k^{\rho g} := x_{k}^g-p_{\rho,k}\).

\emph{Implicit wrap feasibility in IMR.}
IMR replaces \eqref{eq:bmr_partition_bigM} in \eqref{eq:wedge_app} similarly by defining
\begin{align}\label{eq:sigma_app}
\sigma_k := \tfrac12\big(1+\tanh(k_{\mathrm{gate}}\bar s_k)\big)\ \in [0,1] ,
\end{align}
with \(\bar s_k:= n_{x,k}^\top \widehat{r_k^{ag}}_\varepsilon\) and gains $k_{\mathrm{gate}}>0$.



\subsection{Mode-conditioned cable inextensibility}
\label{app:ci_gaps}

This appendix instantiates a gap construction aligned with the mode-gated inextensibility indicators in \eqref{eq:ref_ocp_inext}.
For the binary formulation (BMR), let $\delta_k^{\mathrm r}\in\{0,1\}$ denote the redirected-mode indicator and $\delta_k^{\mathrm d}:=1-\delta_k^{\mathrm r}$.
Define the mode-wise effective lengths $d^{\mathrm d}(z_k)$ and $d^{\mathrm r}(z_k)$ as in \eqref{eq:inst_lengths} (with $p_{\rho,k}$ for the redirected path).
The mode-gated gaps are
\begin{align}\label{eq:ci_gaps_bmr_app}
    g_k^{\mathrm d} &:= \big(L_0+M_{\ell}(1-\delta_k^{\mathrm d})\big)-d^{\mathrm d}(z_k),\nonumber\\
    g_k^{\mathrm r} &:= \big(L_0+M_{\ell}(1-\delta_k^{\mathrm r})\big)-d^{\mathrm r}(z_k),
\end{align}
and the effective gap used in the tensioning implicit constraint is
\begin{align}
g_k := \delta_k^{\mathrm d} g_k^{\mathrm d}+\delta_k^{\mathrm r} g_k^{\mathrm r}.
\label{eq:gap_eff_bmr_app}
\end{align}
This construction ensures that the inactive mode receives a relaxed (large-$M_{\ell}$) gap, preventing the tensioning constraint from being driven by an inactive length map.
The IMR case follows similarly by replacing $(\delta_k^{\mathrm d},\delta_k^{\mathrm r})$ with $(\sigma_k,1-\sigma_k)$, consistent with \eqref{eq:IMR_gamma}.

\subsection{Torque-channel diagnostic used in Fig.~\ref{fig:spatial-state-traj}}
\label{app:tau_diag}

Let $\tau_m(z,1)$ denote the box torque induced by mode $m$ under unit tension, i.e., the $\tau$-component of $w^m(z,1)$ (App.~\ref{app:wrench}). 
Define the redirected--direct unit-tension difference
\(
\Delta\tau_1(z) := \tau_{\mathrm{r}}(z,1)-\tau_{\mathrm{d}}(z,1).
\)

The plotted effective indicator depends on the mode representation:
(i) IMR: $\Delta\tau_{\mathrm{eff}}(z_k) := (1-\sigma_k)\Delta\tau_1(z_k)$, where $\sigma_k$ is the routing gate (App.~\ref{app:feasibility});
(ii) BMR: replace $(1-\sigma_k)$ by the redirected indicator $\delta_k^{\mathrm{r}}$ in \eqref{eq:BMR_binary};
(iii) FMR: use the simplex mixture relative to direct transmission,
\(
\Delta\tau_{\mathrm{eff}}^{\mathrm{FMR}}(z_k)
:= \sum_{m\in\mathcal{M}\setminus\{\mathrm{d}\}}\delta_k^{m}\,\big(\tau_m(z_k,1)-\tau_{\mathrm{d}}(z_k,1)\big),
\)
which is nonzero when multiple mode weights are simultaneously active.

\bibliography{references}  

\end{document}